\documentclass[runningheads]{llncs}
\usepackage{graphicx}

\usepackage{tikz}
\usepackage{comment}
\usepackage{amsmath,amssymb} 
\usepackage{color}

\usepackage{graphicx}
\usepackage{bbm}
\usepackage{bm}
\usepackage{xspace}
\usepackage[utf8]{inputenc} 
\usepackage{url}            
\usepackage{booktabs}       
\usepackage{amsfonts}       
\usepackage{nicefrac}       
\usepackage{microtype}      
\usepackage{algorithm}
\usepackage{algorithmic}
\usepackage{array}
\usepackage{multirow}
\usepackage{enumitem}
\usepackage{makecell}
\usepackage{gensymb}         
\usepackage{mathtools}
\usepackage{dblfloatfix}
\usepackage{pifont}
\usepackage[permil]{overpic}

\usepackage{titletoc,tocloft}

\newcommand{\ba}{{\mathbf{a}}}

\newcommand{\bc}{{\mathbf{c}}}

\newcommand{\bu}{{\mathbf{u}}}

\newcommand{\bx}{{\mathbf{x}}}
\newcommand{\by}{{\mathbf{y}}}
\newcommand{\bz}{{\mathbf{z}}}

\newcommand{\bV}{{\mathbf{V}}}
\newcommand{\bW}{{\mathbf{W}}}

\newcommand{\cI}{{\mathcal{I}}}

\newcommand{\cM}{{\mathcal{M}}}

\newcommand{\cP}{{\mathcal{P}}}
\newcommand{\cS}{{\mathcal{S}}}
\newcommand{\cT}{{\mathcal{T}}}

\newcommand{\cV}{{\mathcal{V}}}

\newcommand{\bbeta}{\bm{\beta}}
\newcommand{\bgamma}{\bm{\gamma}}

\makeatletter
\DeclareRobustCommand\onedot{\futurelet\@let@token\@onedot}
\def\@onedot{\ifx\@let@token.\else.\null\fi\xspace}

\def\eg{\emph{e.g}\onedot} 
\def\ie{\emph{i.e}\onedot}

\def\etal{\emph{et al}\onedot}
\makeatother

\DeclareMathOperator{\Sinkhorn}{SD}

\newcommand{\RT}{\Psi}

\def\argmin{\mathop{\mathrm{arg}\, \mathrm{min}}\limits}

\def\argmin{\mathop{\mathrm{arg}\, \mathrm{min}}\limits}

\usepackage{xcolor}
\usepackage{ifthen}
\usepackage{textcomp}
\definecolor{red}{rgb}{1,0,0}
\definecolor{slateblue}{rgb}{0.7,0.35,0.9}
\definecolor{green}{rgb}{0,1,0}
\definecolor{mahogany}{rgb}{0.75, 0.25, 0.0}
\definecolor{purple}{rgb}{0.6, 0, 0.6}
\definecolor{darkpurple}{rgb}{0.3, 0, 0.3}
\definecolor{darkgreen}{rgb}{0, 0.4, 0}
\definecolor{frenchblue}{rgb}{0.0, 0.45, 0.73}
\definecolor{blue}{rgb}{0,0,1}
\definecolor{goldenrod}{rgb}{0.65, 0.45, 0.03}
\definecolor{gray}{rgb}{0.5,0.5,0.5}
\definecolor{gold}{rgb}{1.0, 0.874, 0}
\definecolor{silver}{rgb}{0.67,0.67,0.67}
\definecolor{brown}{rgb}{0.8, 0.678, 0.4}

\newboolean{revising}
\setboolean{revising}{true}
\ifthenelse{\boolean{revising}}
{
    \newcommand{\ignore}[1]{}

} {
    \newcommand{\ignore}[1]{}

}

\usepackage{cite}
\usepackage[pagebackref,breaklinks,colorlinks]{hyperref}

\usepackage[capitalize]{cleveref}
\crefname{section}{Sec.}{Secs.}
\Crefname{section}{Section}{Sections}
\Crefname{table}{Table}{Tables}
\crefname{table}{Tab.}{Tabs.}

\usepackage[accsupp]{axessibility}  

\begin{document}
\pagestyle{headings}
\mainmatter
\def\ECCVSubNumber{4733}  

\title{Multiview Regenerative Morphing\\ with Dual Flows} 

\titlerunning{Multiview Regenerative Morphing}
%
\author{Chih-Jung Tsai\inst{1} \and
Cheng Sun\inst{1,2} \and
Hwann-Tzong Chen\inst{1,3}}
\authorrunning{C. Tsai et al.}
%
\institute{National Tsing Hua University \and
ASUS AICS Department \and
Aeolus Robotics}
\maketitle
\begin{figure*}
    \centering
    \includegraphics[width=0.99\textwidth]{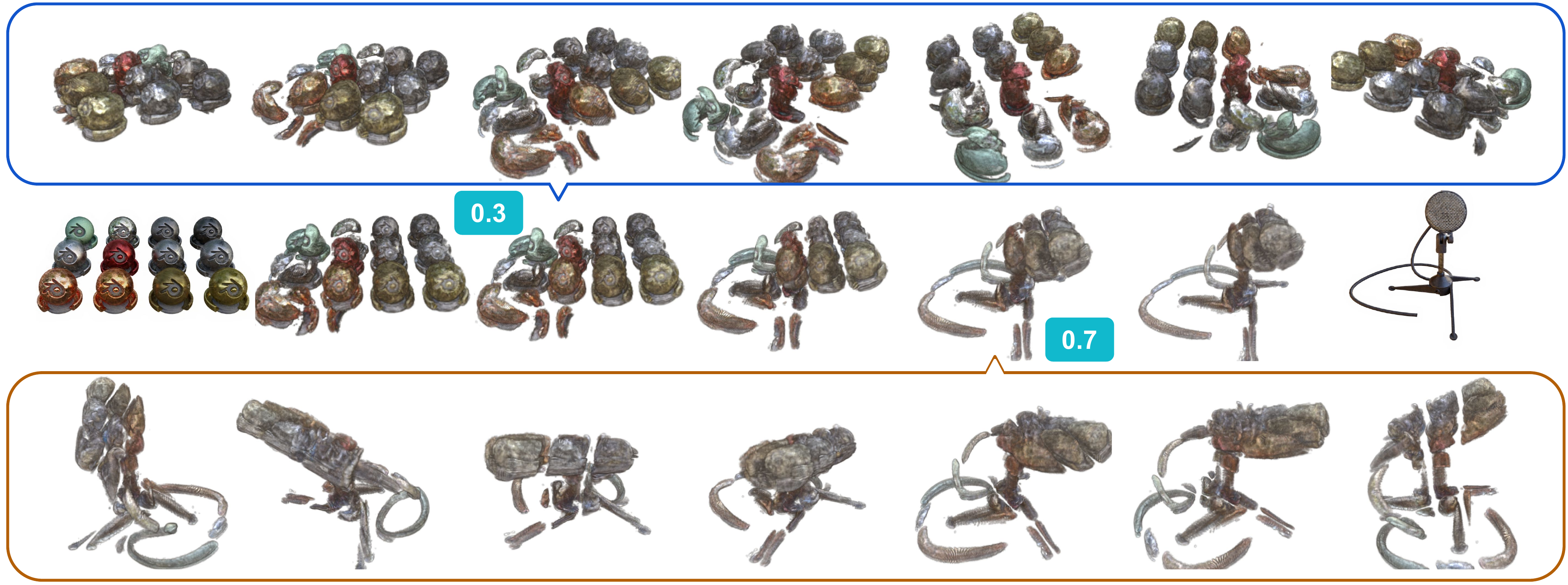}
    \caption{Multiview Regenerative Morphing.}{The middle row shows an example of {\em Multiview Regenerative Morphing} from the source (left) to the target (right). The top row shows multiview rendering when the blending weight is 0.3. The bottom row shows multiview rendering with blending weight 0.7.
    \label{fig:teaser}}
\end{figure*}

\begin{abstract}
This paper aims to address a new task of image morphing under a multiview setting, which takes two sets of multiview images as the input and generates intermediate renderings that not only exhibit smooth transitions between the two input sets but also ensure visual consistency across different views at any transition state. To achieve this goal, we propose a novel approach called Multiview Regenerative Morphing that formulates the morphing process as an optimization to solve for rigid transformation and optimal-transport interpolation. Given the multiview input images of the source and target scenes, we first learn a volumetric representation that models the geometry and appearance for each scene to enable the rendering of novel views. Then, the morphing between the two scenes is obtained by solving optimal transport between the two volumetric representations in Wasserstein metrics. Our approach does not rely on user-specified correspondences or 2D/3D input meshes, and we do not assume any predefined categories of the source and target scenes. The proposed view-consistent interpolation scheme directly works on multiview images to yield a novel and visually plausible effect of multiview free-form morphing. Code: \href{https://github.com/jimtsai23/MorphFlow}{https://github.com/jimtsai23/MorphFlow}
\end{abstract}

\section{Introduction}
\label{sec:intro}
Image morphing is an appealing visual effect that transforms one image into another with coherent intermediate results showing smooth transitions. It has wide applications in visualization, special visual effect, and virtual reality. Conventional morphing methods consist of three steps: \emph{i}) acquire user-specified landmark-based or dense correspondences, \emph{ii}) warp each image into an intermediate layout based on the aligned correspondences, and \emph{iii}) blend the two aligned images with the respective weights. The need of user-specified correspondences is unfavorable and sometimes even impossible when image contents are very dissimilar. On the other hand, as observed in \cite{seitz1996view}, simple warping functions may cause unnatural appearances when characterizing complex deformations. Later approaches like {\em regenerative morphing} \cite{shechtman2010regenerative}, and more recently the GAN-based methods such as \cite{KarrasLA19,KarrasLAHLA20}, are able to relax the requirement of explicitly specified correspondences and produce intriguing effects of image-to-image warping and interpolation.

In this work, we build upon the success of prior techniques and aim to solve a more challenging task of image morphing to render multiview morphs between two structurally unaligned and visually unrelated scenes, as shown in \cref{fig:teaser}. 
More precisely, the new task we seek to address can be described as follows: Consider two sets of images, each taken from a scene of arbitrary categories under various viewpoints. The goal of our task is to build a model that has the capability of producing multiview morphs, such that,  \emph{i}) at any chosen viewpoint, it can generate a sequence of transitions as in standard image morphing, while \emph{ii}) at any given transition moment, it can present multiview renderings of the intermediate morphing scene. To achieve this goal, we propose to learn a model comprising volumetric scene representations for rendering morphs. Each scene is represented by a 4D volume, where each voxel contains RGB and alpha (opacity) values. The representation is learned in a coarse-to-fine manner. First, we use a coarse voxel grid to locate the probable occupancy of the scene, and then we use a finer voxel grid to optimize for the details. Instead of using implicit functions or any kind of neural network, we adopt an explicit differentiable volume rendering scheme to reduce computation time.

Suppose that we have derived the aforementioned representations from the multiview images of the source and target scenes. Now, to proceed with the morphing task, we need to fuse the two representations into an intermediate one that can be used to render novel views of the morphs for any given transition moment. Without relying on predefined correspondences, we adopt an optimal transport mechanism that models intermediate transitional representations as interpolations of the original two representations. 
The interpolations resemble a mixture of scaffolds of the two scenes for querying and blending the volumetric representations. As free-form regenerative morphing may result in shattered structures during transitions, we regularize the morphing process by enforcing a rigid transformation to avoid generating fragmented morphs.   
With the interpolated representation and rigid transformation, our method can render a sequence of coherent morphs of the two category-independent input scenes. The rendered morphs can be displayed from different viewpoints at different transition moments. 

We summarize the main ideas of this work as follows:
\begin{enumerate}[itemsep=1pt,topsep=1pt]
\item This paper presents an optimization-based method that tackles a new task of multiview regenerative morphing as illustrated in \cref{fig:teaser}. The proposed method takes multiview images as the input; no 2D or 3D meshes are needed.
\item Our approach does not assume the categories of and the affinities between the source and target images, nor does it require any predefined correspondences between them. 
\item
Our approach adopts the mechanism of optimal transport to get an interpolated volume for rendering transitional multiview morphs. We also include a rigid transformation in the morphing process to favor `structure-preserving' morphs when possible.
\item
Our method is efficient in learning and rendering. It can learn a morphing renderer from scratch (directly from the input images) in 30 minutes. For morphing and rendering, the learned renderer can generate one novel-view morph per second.
\end{enumerate}

\section{Related Work}
\paragraph{Image morphing.}
Image morphing aims at transforming a source image to a target image smoothly and with natural-looking in-between results.
Traditional approaches~\cite{wolberg1998image,lerios1995feature} use image warping and color interpolation with predefined dense correspondence.
In particular, \cite{alexa2000rigid} enforces the transition to be as rigid as possible, and \cite{seitz1996view} considers camera viewpoint to prevent distortion.
Patch-based methods~\cite{shechtman2010regenerative,darabi2012image} are later proposed to synthesize in-between images using source and target patches under temporal coherence constraints.
Recently, Generative Adversarial Networks (GAN) has shown impressive image generation results by learning a projection from latent space to image space.
Image morphing can then be achieved by simple linear blending in the GAN latent space~\cite{WuNSL21,PanZDLLL20,AbdalQW19,FishZPCSB20}.
Simon and Aberdam~\cite{SimonA20} further propose to solve the Wasserstein barycenter problem constrained on GAN latent space to achieve smooth transitions and natural-looking in-between results.
Optimal transport has also been used to produce morphing between simple 2D geometries \cite{bonneel2011displacement,bonneel2016wasserstein,solomon2015convolutional,benamou2015iterative}. However, the morphs lack textures as in nature images. Our method is different from the above morphing methods in that we generate 3D representations and render view-consistent morphs in novel views. Perhaps the most similar to us are the multilevel free-form deformation morphing techniques described in \cite{wolberg1998image}. While they still depend on 3D primitives and human-labeled correspondence, our morphing technique is fully automatic and unsupervised.

\paragraph{Volume renderer from multiview images.}
Reconstructing a volumetric scene representation that supports novel-view synthesis from a set of images is a long-standing task with steady progress \cite{de1999poxels,szeliski1999stereo}.
NeRF~\cite{MildenhallSTBRN20} has recently revolutionized this task by incorporating the coordinate-based multilayer perceptrons (MLP) to represent each spatial point's color and volume density implicitly.
The MLP model is trained to minimize the photometric loss on the observed views with differentiable volume rendering.
Many follow-up works of NeRF are proposed to achieve better qualities on background~\cite{ZhangRSK20}, surface~\cite{OechslePG21}, multi-resolution~\cite{BarronMTHMS2021}, imperfect input poses~\cite{JeongACACP21,LinCTL21,MengCLWSXHY21}, fewer input views~\cite{Yu20arxiv_pixelNeRF,ChenXZZXYS21,WangWGSZBMSF21}, and dynamic scene~\cite{GaoSKH21,Li20arxiv_nsff,Martin-BruallaR21,Xian20arxiv_stnif}.
Despite the high quality and flexibility, NeRF still has a disadvantage of its lengthy training and rendering run-time.
To improve rendering speed, many methods are proposed to convert the trained implicit MLP representations to explicit voxel-grid or hybrid representations~\cite{WizadwongsaPYS21,GarbinKJSV2021,YuLTLNK2021,HedmanSMBD2021}.
The improvement on training run-time relies on cross-scene pre-training~\cite{Yu20arxiv_pixelNeRF,ChenXZZXYS21,WangWGSZBMSF21} or external depth~\cite{LiuPLWWTZW21,DengLZR21}.
Our morphing algorithm is agnostic to the underlying scene representation reconstruction techniques. For ease of use, our representation explicitly models the scene with voxels, similar to DirectVoxGO \cite{sun2021direct}. We reconstruct the source and target scene representations from their respective image sets, and use the learned volumetric representations for morphing.

\paragraph{Shape interpolation.}
Given two (or more) shapes, shape interpolation aims at generating their in-between shape specifying by a composition percentage, which enables a smooth deformation from one shape to the other.
Traditional methods try to recover the shape space manifold~\cite{KilianMP07,WirthBRS11,HeerenRWW12,HeerenRSWW16}, and then the shape interpolation becomes a geodesic-path searching problem on the manifold.
As shape manifold recovering is challenging, some recent approaches~\cite{EisenbergerLC19,EisenbergerC20} directly find the deformation field from source to target shapes but with isometric (zero-divergence, constant volume) assumption.
NeuroMorph~\cite{eisenberger2021neuromorph} uses neural networks to predict the correspondences and the deformation field, which works well even for non-isometric pairs.
Generative neural networks have recently achieved good results in 3D by building shape latent spaces and using deep decoders to map latent codes to voxels~\cite{WuZXFT16}, signed distance fields~\cite{JiangM17,ParkFSNL19}, point clouds~\cite{CosmoNHKR20,LiZZPS19,ShuPK19}, or meshes~\cite{ChengBZKPZ19}; shape interpolation is then achieved by linearly blending the latent codes of the two shapes.
These shape-interpolation techniques typically take 3D as input (\eg, mesh), and GAN-based methods further require large datasets, while our method only needs two sets of images that capture the source and the target shapes.
Janati \etal and Solomon \etal have considered interpolations between shapes as Wasserstein barycenters \cite{janati2020debiased,solomon2015convolutional}. However, their computations do not include morph in appearances. To facilitate learning on more complicated shapes, we take a different strategy and morph the scenes with the Wasserstein flow \cite{feydy2019interpolating}. We further regularize the flow with rigidity constraints, and use these local and global dual flows to achieve multiview regenerative morphing.

\begin{figure*}
    \centering
    \includegraphics[width=\textwidth]{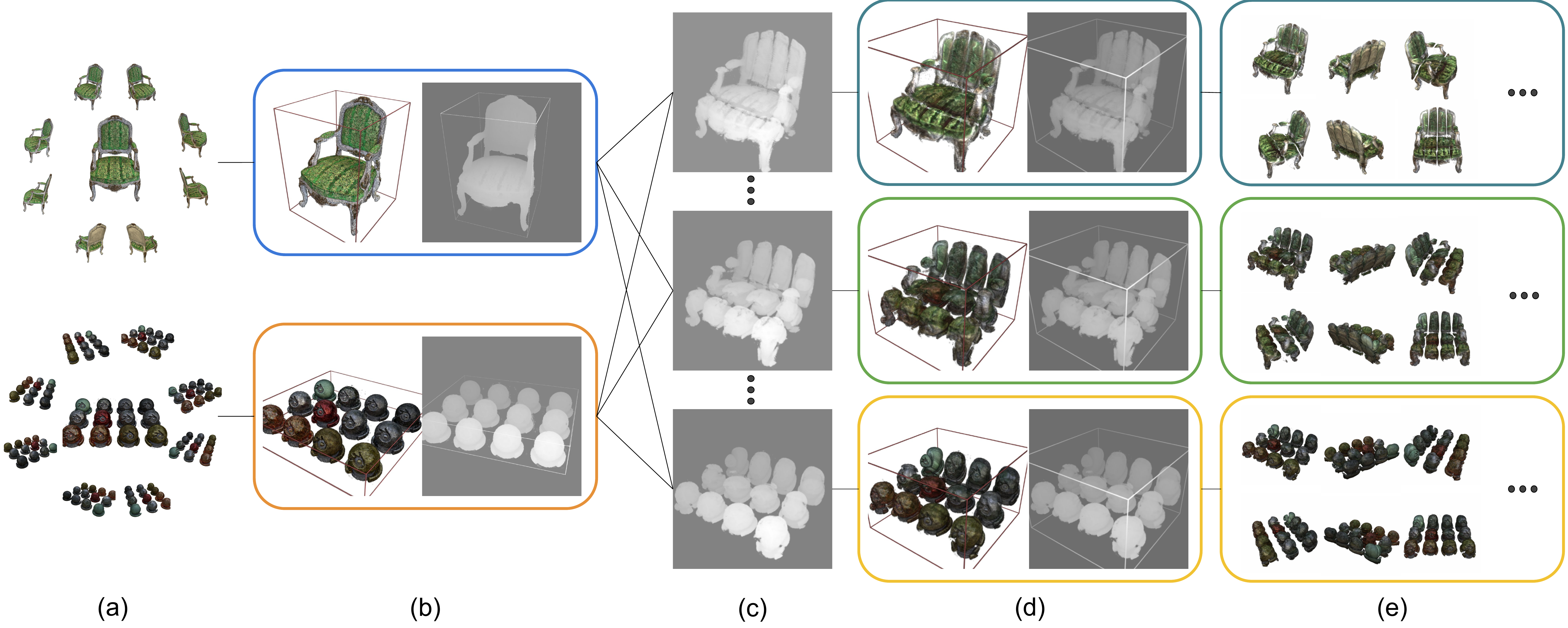}
    \caption{An overview of our method. (a) The input multiview images of the source and target scenes. (b) The volumetric representation of each scene comprises color and opacity information. (c) A sequence of the morphed point sets with different blending weights. They are interpolations between source and target scene in Wasserstein metrics. (d) The volumetric representations generated by morphing. (e) Examples of multiview morphing rendered at arbitrary viewing directions. It can be seen that the rendering results are view-consistent---at any moment, the intermediate morph can be viewed as an actual scene and does not exhibit any conflicts across views.}
    \label{fig:pipeline}
\end{figure*}

\section{Overview}
\label{sec: overview}
Consider the images $\cI^\cS$ and $\cI^\cT$ collected from the source and target scenes with camera poses $\zeta^\cS$ and $\zeta^\cT$. Our method can generate morphed images between $\cI^\cS$ and $\cI^\cT$ given arbitrary weights $t$ and viewing angles $\theta, \phi$.
The method has two phases. In the first phase, we establish volumetric representations for each scene with the purpose of generating view-consistent morphs from unaligned images. The representation comprises opacity and color. In the second phase, we use optimal transport to generate morphs between the derived volumetric representations from the first phase. The morphing process is controlled by rigid transformation (RT) flow and optimal transport (OT) flow. RT flow preserves the impression of the source scene during morphing, preventing shattered generation. OT flow deforms the source scene into the target without the need of correspondences. By applying the two flows, we obtain a morphed representation and render view-consistent morphs in any views. \cref{fig:pipeline} illustrates the pipeline of our method.

Formally, we use a volume $\cV$ to model a scene by mapping a 3D position $\bx = (x,y,z)$ to its corresponding opacity $\alpha$ and color $\bc = (r,g,b)$ as
\begin{equation} \label{eq:voxel_grid_rep}
    \cV: \mathbb{R}^{3} \rightarrow \mathbb{R}^{4}, \quad \cV(\bx)=\left(\cV_\alpha(\bx),\cV_\bc(\bx)\right)\,,
\end{equation}
where $\cV_\alpha(\bx)$ retrieves the opacity $\alpha$ and $\cV_\bc(\bx)$ yields the color $\bc$.
Based on \cref{eq:voxel_grid_rep} we build volumes $\cV^\cS$ and $\cV^\cT$ for representing the source and target scenes.
The opacity can be used to filter out negligible voxels. The balance between granularity and efficiency is controlled by a threshold $\delta_{\alpha}$, \ie, voxel $v_i$ in $\cV_\alpha$ is collected if $\alpha_i > \delta_\alpha$. 
The collection of points and opacity values  serves as a shape representation, which can be  expressed as a weighted point set $\cP = \{ (\omega_{i}, \mathbf{x}_i) \}_{i=1}^{N}$, where $\omega_i = \alpha_i / \sum_j^N \alpha_j$, associating each point with a weight derived from the opacity by normalization.
In this way, we create a source shape $\cS$ from the source volume $\cV^\cS$ and a target shape $\cT$ from the target volume $\cV^\cT$, where both are in the form of a weighted point set as described above. While we transform the opacity volume into the source shape $\cV^\cS$, for each point $i$ in the source set, its color $\bc_i$ can be gathered from the color volume $\cV^\cS_c$. The point colors are preserved for blending the final appearances.

\begin{figure*}
    \includegraphics[width=\textwidth]{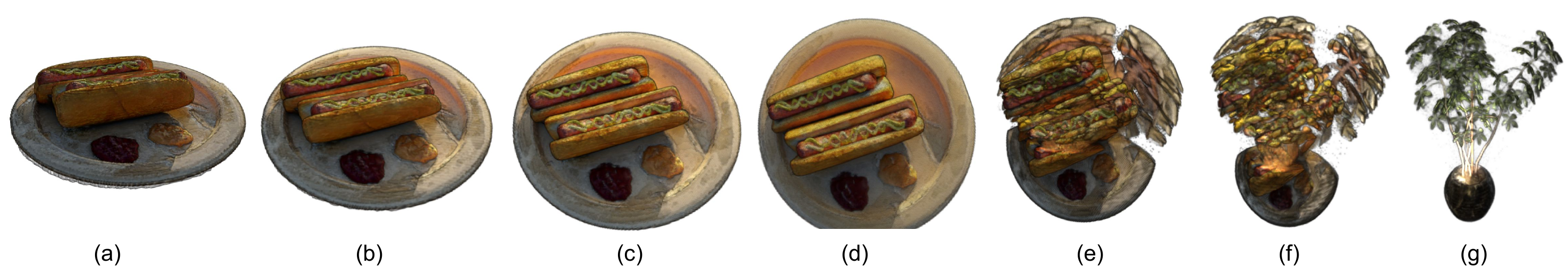}
    \caption{To visualize the different effects of the rigid transformation (RT) flow $\mathbf{f}$ and the optimal transport (OT) flow $\mathbf{g}$, we deliberately apply them one after the other. (a--d): we apply only $\mathbf{f}$, which aligns the source's pose with the target's without changing the shape. (d--g): we apply only $\mathbf{g}$, which deforms from aligned source shape to target shape. In our method, the two flows are jointly applied during the entire morphing process.}
    \label{fig:flow}
\end{figure*}

To morph between the source shape $\cS$ and the target shape $\cT$, we develop a two-step algorithm using the {\em 2--Wasserstein distance}. In the first step, we seek a rigid transformation $\hat{\RT}$ that minimizes the distance between the source and target shapes in the Wasserstein space. We solve for $\hat{\RT}$ using gradient descent on unbiased Sinkhorn divergence $\Sinkhorn$. A more detailed algorithm is described in \cref{ssec:RT_flow}. The second step finds an interpolation between the rigidly transformed source shape $\hat{\RT} (\cS)$ and the target shape $\cT$. 
We obtain the interpolation via a gradient-guided step on $\Sinkhorn$ \cite{genevay2018learning,feydy2019interpolating}, which is interpreted as a displacement vector from the source shape to the target shape. As a result, two flows are created, namely, the RT flow $\mathbf{f}$ and the OT flow $\mathbf{g}$, by comparing the transformation $\hat{\RT} (\cS)$ to the source shape and the target shape to the transformed source. A morphed shape can then be generated by 
\begin{equation}
\label{eq:flow}
    \cM_t \leftarrow \cS +  \mathbf{f}(t) + \mathbf{g}(t) \,,
\end{equation}
where the flow parameter $t \in [0, 1]$ controls the progression of $\mathbf{f}$ and $\mathbf{g}$ at each transition moment $t$ to create the expected morphs. \cref{fig:flow} visualizes the effects of the two flows. 

We use the morphed shape $\cM_t$ to query color volume of target $\cV_\mathbf{c}^\cT$ and generate blended colors $\cV_\mathbf{c}^{\cM_t}$ along with colors of source points. The morphed shape is voxelized to form $\cV^{\cM_t} = \left( \cV_{\alpha}^{\cM_t}, \cV_{\mathbf{c}}^{\cM_t} \right)$ so that we can render view-consistent morphing images using $\cV^{\cM_t}$ under arbitrary viewing directions.

In short, the proposed Multiview Regenerative Morphing provides an on-the-fly morphing renderer trained on two different scenes without any correspondences. The main idea is to extend image morphing from single-view to multiview and generate view-consistent multiview morphs. We introduce an efficient learning strategy in \cref{sec:volume_rendering} to derive volume representations from multiview images. We use optimal transport in 2--Wasserstein space to merge the volumes of the two scenes. The algorithm for computing the morphs is detailed in \cref{sec:barycentric_morphing}.

\section{Volume Renderer}
\label{sec:volume_rendering}
We lift the image morphing problem from single-view to multiview by learning volumetric representations for the source and target scenes. The representation contains shape and appearance, and is learned from multiview images and their camera poses.
Below, we briefly introduce how to reconstruct such a scene representation from the calibrated input images and our design choices.

To obtain volumes as in \cref{eq:voxel_grid_rep}, we adopt differentiable volume rendering to optimize the opacity and color volumes for each scene. Given the image poses $\zeta^\cS$ and $\zeta^\cT$, we assume a pinhole camera and generate rays emitted from the camera center, based on each pixel's position. During rendering, points are sampled along a ray and queried with the scene representation to produce a series of colors and volume densities.
The densities are converted into alpha values via $\alpha_i = 1-\exp(-\sigma_i\delta_i)$ for the follow-up alpha compositing to accumulate the point queries into a single ray color:
\begin{equation}
\label{eq:ray}
\begin{aligned}
    \hat{C}(\mathbf{r}) = \sum_{i=1}^{N} T_{i}\alpha_i\mathbf{c}_{i} ~, \quad
    T_{i} = \prod_{j=1}^{i-1} (1-\alpha_j) \,,
\end{aligned}
\end{equation}
where $\mathbf{r}$ is the camera ray on which the $N$ discrete points are sampled, $T_i$ is the accumulated transmittance from ray emission to the current sample $i$, and $\delta_i$ is the distance between adjacent samples.
The scene representation is optimized by minimizing the photometric mean squared error
\begin{equation}
    \mathcal{L}=\frac{1}{K} \sum_{m=1}^{K} \left\| \hat{C}(\mathbf{r}_m)-C(\mathbf{r}_m) \right\|_{2}^{2} ~,
\end{equation}
where $K$ is the mini-batch size, $\hat{C}$ is the rendered color, and $C$ is the observed pixel color.

There are two common volumetric representations: \emph{i}) voxel grids, which explicitly parameterize the 3D scene as grid values, and \emph{ii}) multilayer perceptrons (MLP), which implicitly learn the mapping via MLP weights. We opt to use the explicit voxel grid to model the scene for faster convergence and for the convenience of latter usage in morphing. During training, we learn volumes of opacity $\cV_\alpha(\bx)$ and color $\cV_\bc(\bx)$ for each scene, while each sample on the rays is trilinearly interpolated with neighboring voxels. We note, however, that the trained implicit representations~\cite{YuLTLNK2021,GarbinKJSV2021} can easily be turned into volumetric representations and used with our morphing algorithm.

\section{Wasserstein Morphing Flow}
\label{sec:barycentric_morphing}
With learned volumetric representations $\cV^\cS$ and $\cV^\cT$, we develop a differentiable morphing algorithm that welds two volumes into a morphed volume for rendering. Since the source and target scenes are not constrained to be in one category and may be very dissimilar, optimal transport is used to deform the source scene into the target. Specifically, we use Sinkhorn divergence (SD), a regularized optimal transport objective that minimizes 2--Wassertein distance between two point sets. The source and target shapes $\cS$ and $\cT$ are created as weighted point sets collected from the volumes $\cV^\cS$ and $\cV^\cT$.
Note that our method assumes the two weighted point sets to be positive discrete measures such that they can be compared in Wasserstein metrics.
Therefore, the weights of the point sets have been normalized to make them discrete probability distributions, \ie, the weights $\{\omega_i^\cS\}_{i=1}^{N^\cS}$ and $\{\omega_j^\cT\}_{j=1}^{N^\cT}$  of the source and target shapes satisfy $\sum_{i}^{N^\cS} \omega_i^\cS = 1$ and $\sum_{j}^{N^\cT} \omega_j^\cT = 1$.  
More precisely, such a discrete measure can be expressed as a sum of weighted Dirac mass, and we thus have $\cS = \sum_{i=1}^{N^\cS} \omega_i^\cS  \, \Delta_{\bx_i^\cS}$ and  $\cT = \sum_{j=1}^{N^\cT} \omega_j^\cT  \, \Delta_{\bx_j^\cT}$, where $\Delta$ is the Dirac delta function that can be thought of as an indicator of occupancy at a given point $\bx$.

Our aim now is to obtain a 3D morphing renderer, where at the core is a morphed volumetric representation $\cV^{\cM_t}$. 
We view the morphing process as solving an optimal transport problem for moving mass from a source distribution to a target distribution. To generate smooth transitions, we design a flow-based morphing scheme, which comprises the rigid transformation (RT) flow and optimal transport (OT) flow. The RT flow pushes the source scene toward the target scene by applying rotation and translation. As the rigid transformation is global, it can preserve the original appearance. On the other hand, the OT flow provides smooth local deformations but may change the topology of the shape. In what follows, we first describe how to compute the RT flow for globally registering the two shapes, and then we detail the algorithm of OT flow, as well as the complete scheme of dual-flow based morphing.

\subsection{Rigid transformation flow}
\label{ssec:RT_flow}
With the two shapes $\cS$ and $\cT$ expressed as two discrete measures that are derived from the source and target volumes, we estimate the rigid transformation $\RT \in \mathrm{SE(3)}$ by minimizing the Sinkhorn divergence $\Sinkhorn$ \cite{feydy2019interpolating} between the transformed source measure and the target measure:
\begin{equation}
\label{eq:rigid_transformation}
    \hat{\RT} = \argmin_{\RT} \; \Sinkhorn\left( \RT(\cS), \cT \right) \,,
\end{equation}
where the transformed measure $\RT(\cS)$ is defined in the form of weighted sum of Dirac mass by 
\begin{equation} \RT(\cS) = \sum_{i=1}^{N^\cS} \omega_i^\cS \, \Delta_{\RT (\mathbf{x}_i^\cS)} \,.
\end{equation}
The rigid transformation $\RT$ comprises rotation $\mathbf{R}$ and translation $\mathbf{z}$. The initial states $\mathbf{R}^{(0)} = \mathbf{I}_3$ and $\mathbf{z}^{(0)} = \mathbf{0}$ are updated by the gradients $\nabla_{\mathbf{R}} \Sinkhorn (\RT(\cS), \cT)$ and $\nabla_{\mathbf{z}} \Sinkhorn (\RT(\cS), \cT)$. Note that directly applying the gradient update to $\mathbf{R}$ may lead to an unconstrained projection matrix, and therefore we replace the gradient-updated matrix with identity singular values via singular value decomposition (SVD). Specifically, we compute
\begin{equation}
     \left(\mathbf{R} -\nabla_{\mathbf{R}} \Sinkhorn(\RT(\cS), \cT) \right) = \mathbf{U} \mathbf{\Sigma} \bV^{\intercal} \,,
\end{equation}
and use $\mathbf{U} \bV^{\intercal}$ as a surrogate for the new rotation matrix. 
Finally, we can solve \cref{eq:rigid_transformation} for the estimated transformation $\hat{\RT}$, and the transformation flow $\mathbf{f}$ parameterized by the time step $t \in [0,1]$ is then given by 
\begin{equation}
\label{eq:transformation_flow}
    \mathbf{f}(t) = t \cdot (\hat{\RT}(\cS) -  \cS) \,,
\end{equation}
which is used in \cref{eq:flow} to provide the progression on the source shape $\cS$.
For brevity, the definition of Sinkhorn divergence $\Sinkhorn$ and the derivation of the gradients $\nabla_{\mathbf{R}}\Sinkhorn$ and $\nabla_{\mathbf{z}} \Sinkhorn$ are omitted here. More details can be found in the supplementary material.

\subsection{Optimal transport flow}
\label{ssec:OT_flow}
The rigid transformation makes the two measures $\hat{\RT}(\cS)$ and $\cT$ distribute in similar loci in $\mathbb{R}^3$. We can further find a smooth deformation between them using optimal transport. In our method, the interpolation is achieved by adding the gradient that minimizes the Sinkhorn divergence between the transformed shape and the target. Given a time step $t \in [0,1]$ as a blending weight, the optimal transport flow can be written as
\begin{equation}
    \mathbf{g}(t) = - t \cdot \nabla_{\mathbf{x}} \Sinkhorn(\hat{\RT}(\cS), \cT) \,,
\end{equation}
\noindent
which facilitates on-the-fly rendering with a varying time step $t$. Instead of applying the two flows $\mathbf{f}$ and $\mathbf{g}$ sequentially, we advance the two flows simultaneously on $t \in [0, 1]$ as shown in \cref{eq:flow}, and get the morphed measure $\cM_t$ as the morphed shape. The disentanglement of the two flows in Wasserstein metrics enables the morphing to evolve as rigidly as possible even under inevitable topology changes.

We voxelize the morphed measure $\cM_t$ into $\cV_\alpha^{\cM_t}$ by collecting the points with histograms. Each point allocates its weight to the bin according to its position. The histogram is then transformed to a volume. The discretization has to be implemented with care to prevent aliasing. Here we first spread each point in $\cM_t$ to its eight neighbors on the grid. The weight also splits into eight according to the distances between points. The histogram gathers weights of all points and generates $\cV_\alpha^{\cM_t}$. $\hat{\cM}_{t=1}$ is used to query the color volume $\cV_\mathbf{c}^\cT$ for the corresponding target colors. The color of each morphed point is blended between the source color and target color. The morphed color volume $\cV_\mathbf{c}^{\cM_t}$ is generated using a histogram method similar to $\cV_\alpha^{\cM_t}$.


\section{Results}
We evaluate our method on real and synthetic datasets, including \textbf{Synthetic--NeRF} \cite{Mildenhall20eccv_nerf}, \textbf{Synthetic--NSVF} \cite{LiuGLCT20}, \textbf{Tanks\&Temples} \cite{knapitsch2017tanks} and \textbf{BlendedMVS} \cite{yao2020blendedmvs}. Each datasets contains scenes with surrounded imaging and their camera poses. 

\subsection{Multiview regenerative morphing}

\begin{figure*}
    \centering
    \includegraphics[width=0.95\textwidth]{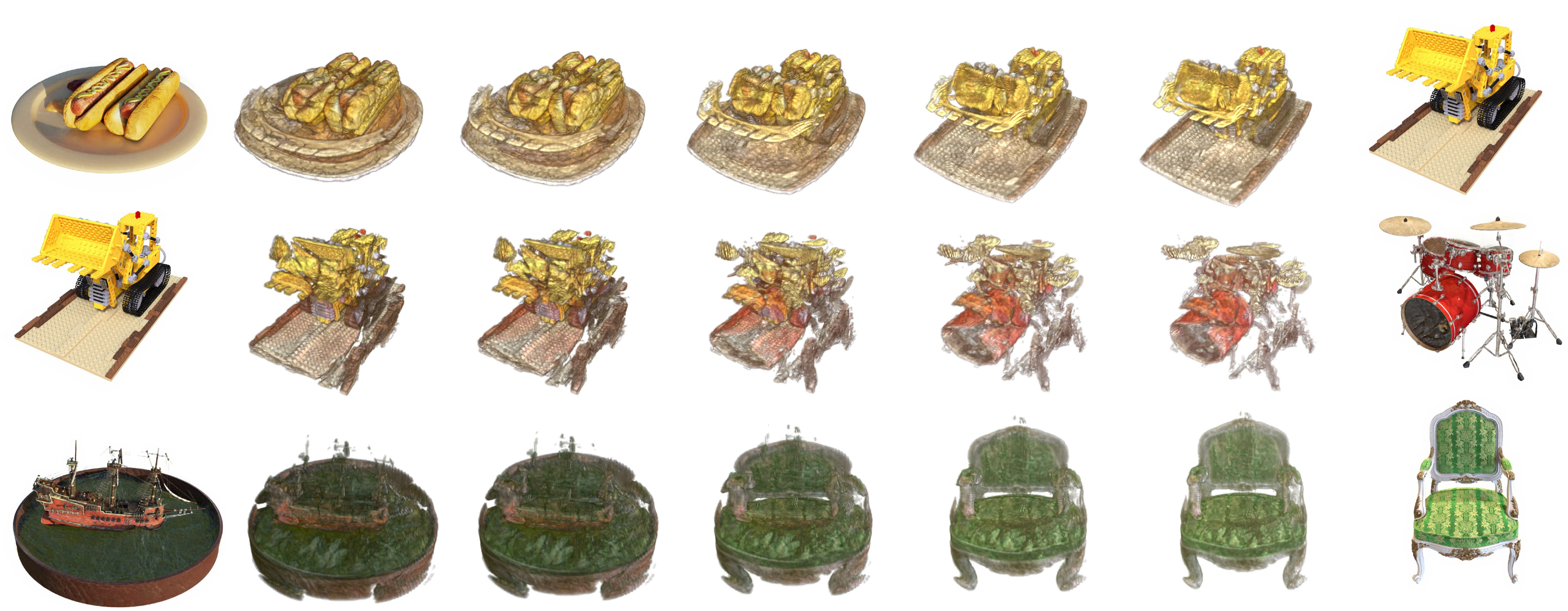}
    \caption{Morphing of scenes in Synthetic--NeRF. Each row shows a smooth transition from the source (left) to the target (right).}
    \label{fig:nerf}
\end{figure*}

Each scene may contain a number of objects, where the morphing between scenes needs to divide or merge objects smoothly. \cref{fig:teaser} shows morphed images from materials with different colors and reflections into a single microphone.
The top and bottom rows show the different views of the morph frozen at $t=0.3$ and $t=0.7$. The rendering results are view-consistent, \ie, at the frozen moment, the intermediate morph can be viewed as an actual coherent scene and exhibits no conflicts across different views.

\begin{figure*}
    \centering
    \includegraphics[width=0.97\textwidth]{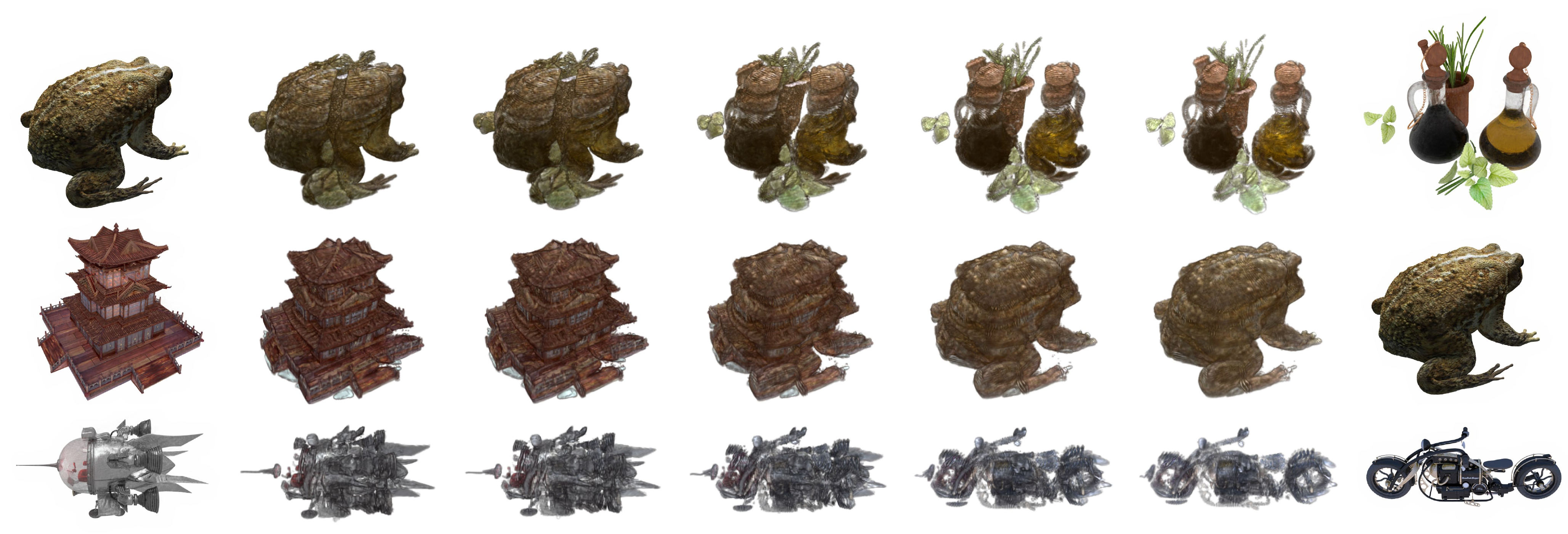}
    \caption{Morphing of scenes in Synthetic--NSVF. Each row shows a smooth transition from the source (left) to the target (right).}
    \label{fig:nsvf}
\end{figure*}

\cref{fig:nerf} shows morphing with scenes in Synthetic-NeRF. We demonstrate smooth transitions between three different sets of source and target scenes. Especially, in the middle row a lego truck morphs into a drum set, with very complex detail. In \cref{fig:nsvf}, we show morphing results in Synthetic-NSVF. 
Due to the limitation of space, we provide more results of multiview rendering in videos in the supplementary material.


\begin{figure*}
    \centering
    \includegraphics[width=\textwidth]{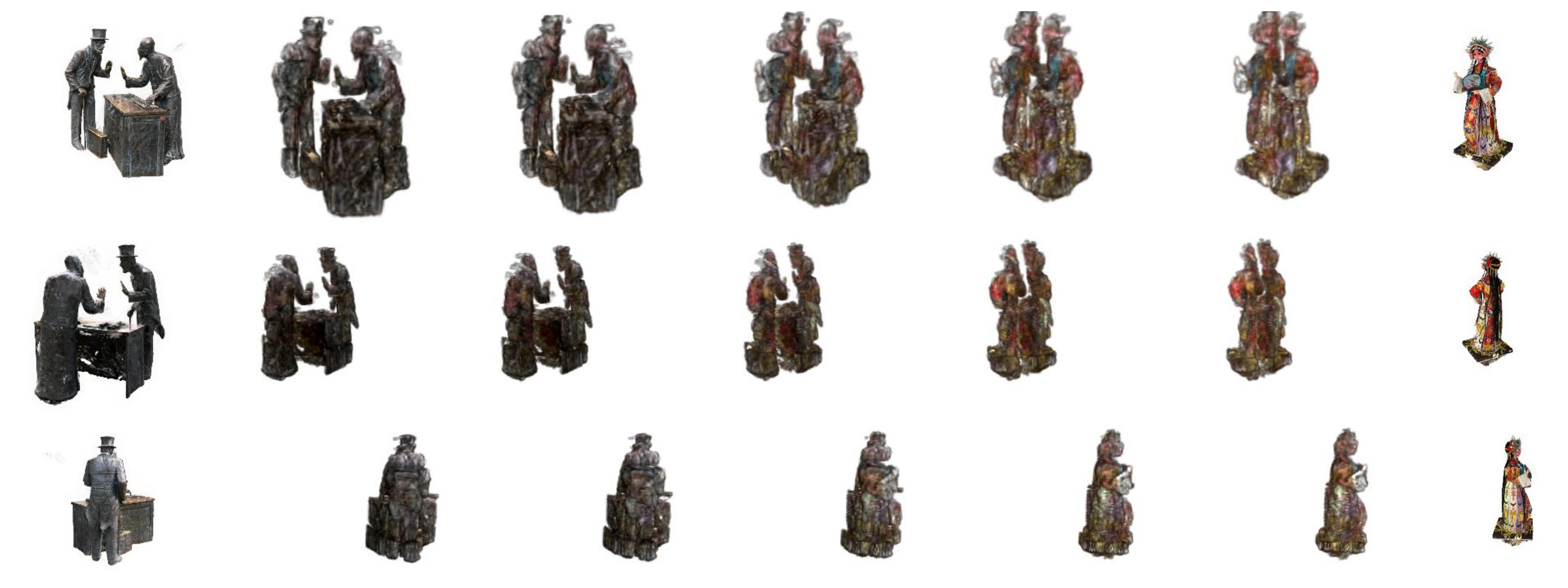}
    \caption{Morphing of scenes in BlendedMVS. All the rows exhibit the same transition, while each row shows renderings under some view.}
    \label{fig:mvs}
\end{figure*}

\cref{fig:mvs} demonstrates the morphing between `Statue' and `Character'. Both scenes are from BlendedMVS. The three rows represent the same transition rendered in different viewpoints. Two people in the Statue gradually get close to each other and merge into a single person. Also the scene becomes colorful, from metallic texture into custom and makeup. Since the training images have abundant specular lighting, resulting in noises and heavy shadows, we therefore use thresholding and color distribution manipulation on the morphed volumes to compensate the flaw.

\begin{figure*}
    \centering
    \includegraphics[width=\textwidth]{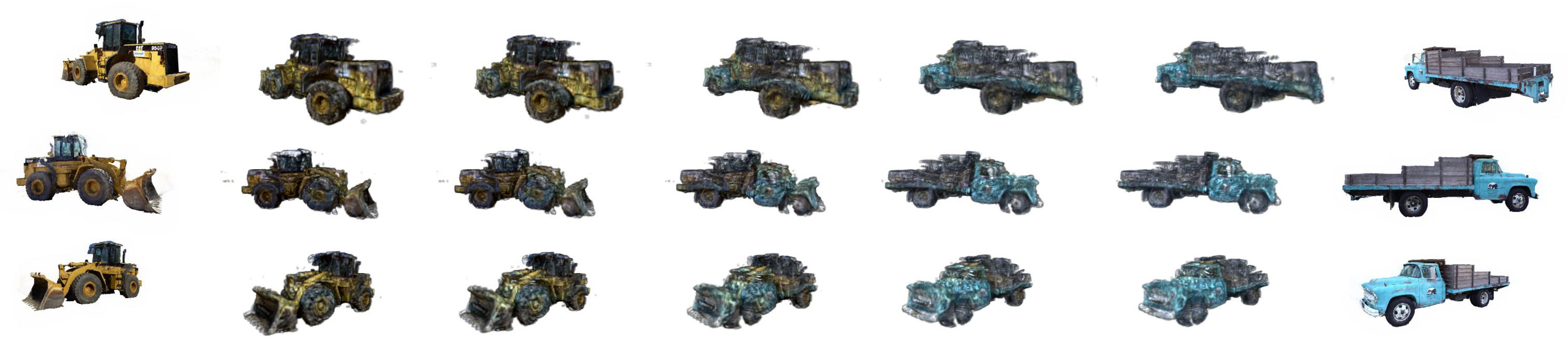}
    \caption{Morphing of scenes in Tanks\&Temples. All the rows exhibit the same transition, but in different views.}
    \label{fig:tt}
\end{figure*}

\begin{figure*}
    \centering
    \includegraphics[width=\textwidth]{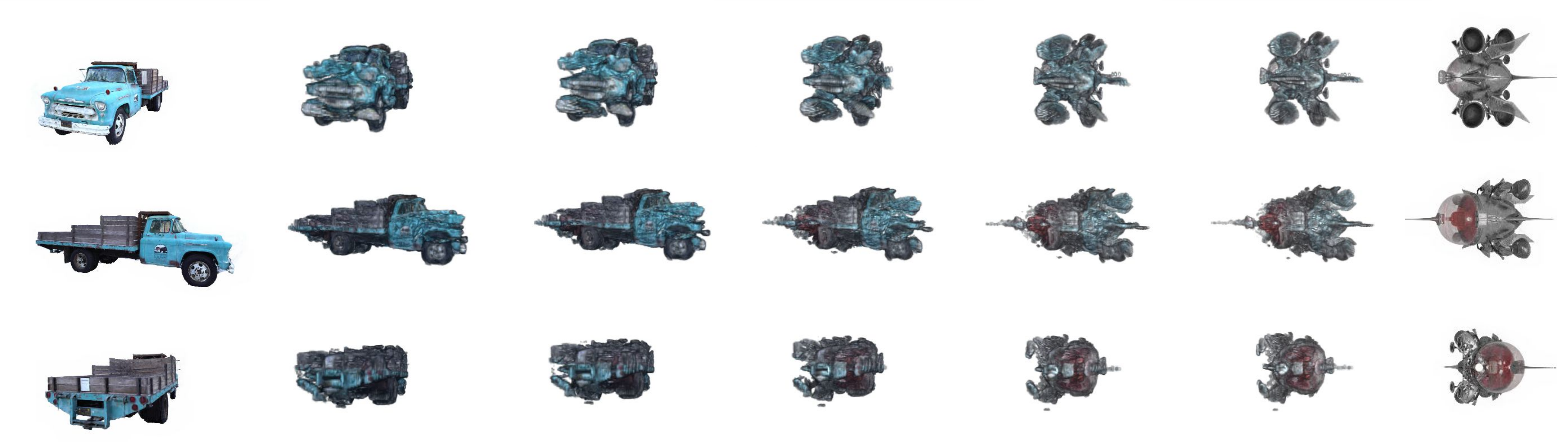}
    \caption{Morphing between real and synthetic scenes. We demonstrate view-consistent morphing by rendering in three different views.}
    \label{fig:truck2ship}
\end{figure*}

\begin{figure*}
    \centering
    \includegraphics[width=0.95\textwidth]{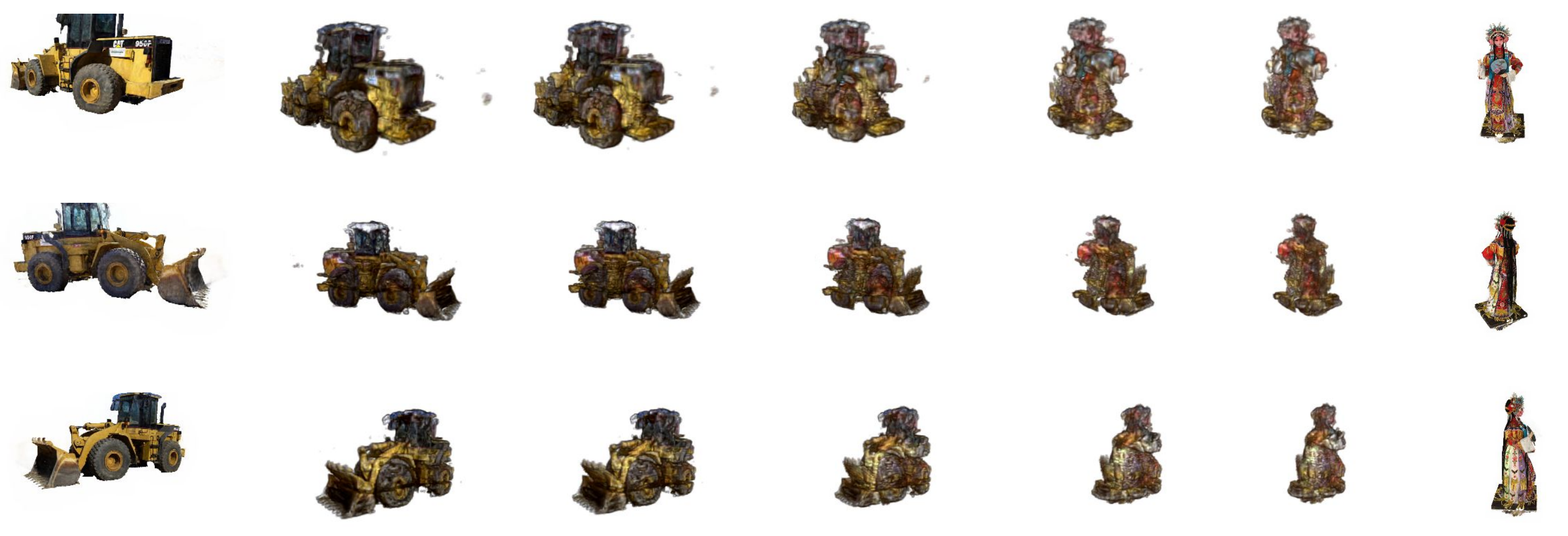}
    \caption{Morphing between real and synthetic scenes. The morphs are rendered with three viewpoints, one in each row.}
    \label{fig:cat2character}
\end{figure*}

\cref{fig:tt} shows transitions between `Caterpillar' and `Truck'. Both scenes are from Tanks\&Temples. The two shapes are reconstructed with collected images, under varying lighting conditions. The direct result has the floating noise around the shapes. We use simple thresholding to remove the noise in the space of the volume. Each row shows renderings of the transition under some view. The morphs are view-consistent in each column. Each column relates to some blending weight.
In addition to morphing real scenes, we demonstrate morphing between real and synthetic scenes. \cref{fig:truck2ship} shows the morphs from a real truck to synthetic `Spaceship'.  Likewise, \cref{fig:cat2character} interpolates between `Caterpillar' and `Character' from BlendedMVS. The real to synthetic scene morphing is achieved under our normal setting.

\paragraph{Limitations on real scenes.}
Our dual-flow multiview morphing is a model-free method and therefore not restricted to specific data domains. \cref{fig:tt} shows the morphing between two real scenes, where our method generates reasonable transitions between the two very dissimilar scenes. However, due to the lighting changes across views, the rendered morphs contain more noise in comparison with synthetic data. The quality of multiview morphing relies on the reconstructed volumetric representations of the original scenes. Those reconstruction artifacts tend to remain during the entire morphing process. Similar artifacts are observed in \cref{fig:mvs}, where the reflection of the surface affects the reconstruction.

\subsection{Comparisons with other approaches}
We validate that our morph generation is geometry-aware in contrast to other correspondence-free morphing algorithms. We compare our method with a 3D-based method, Debiased Sinkhorn \cite{janati2020debiased}, and a 2D-based method, Deep Image Analogy \cite{liao2017visual}. To compare with the 2D-based method, we use the optimized volume of each scene to generate pose-aligned images as their input. NeRF's eight scenes are used for evaluation. We randomly sample camera poses from the upper hemisphere for different scenes and render the morphs with varying transition weights. For each transition, we use COLMAP to solve {\em structure from motion}. As a result, $83.3\%$ of the morphing images generated by our method are successfully registered by COLMAP, which means our method can mostly render 3D consistent morphs. 
On the other hand, Debiased Sinkhorn \cite{janati2020debiased} generates blur images, resulting in poor reconstruction: Only $16.7\%$ of its morphing images can be successfully registered by COLMAP. Deep Image Analogy \cite{liao2017visual} generates visually pleasing morphs, but it is not robust to view changes, as mentioned in \cite{seitz1996view}: not surprisingly, no consistent 3D structure can be reconstructed.

\subsection{Ablation study}
We evaluate different aspects of our method, especially the effect of the RT flow and the OT flow. We compare the rendering of direct optimal-transport morphing with our rigid-transformation-enabled OT morphing. We also demonstrate the effect when only one of the RT flow or the OT flow is applied.

\begin{figure*}
    \includegraphics[width=\textwidth]{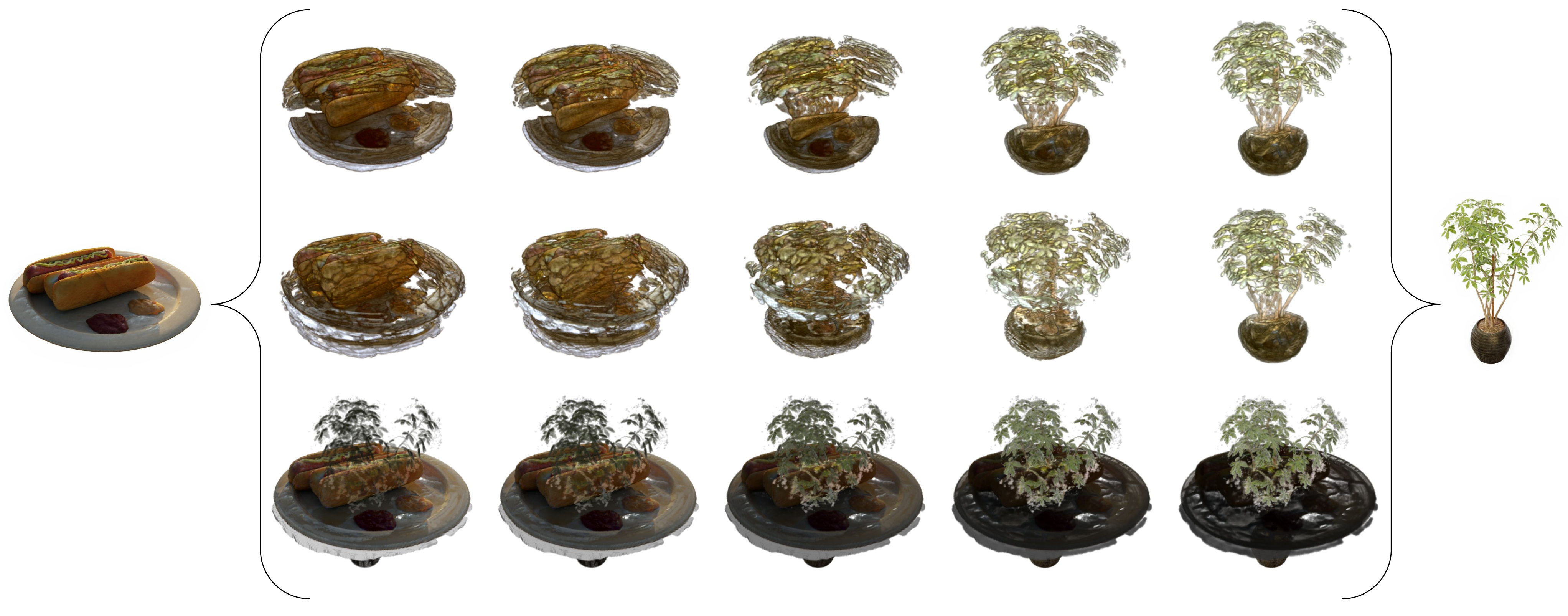}
    \caption{Comparisons between three morphing method. First row shows our dual-flow morphing. Second row uses only OT flow. Third row use no flow, but simple blending.}
    \label{fig:6d}
\end{figure*}

\paragraph{Rigid transformation flow:}
As mentioned in \cite{schaefer2006image,alexa2000rigid,eisenberger2021neuromorph}, achieving as-rigid-as-possible transformation is an appealing property for morphing. The property preserves the original shape during transition and thus helps to produce plausible intermediate results. Unlike previous methods that generate meshes for deformation, here we formulate the rigid transformation as a flow in Wasserstein space. 
\cref{fig:6d} shows comparisons of transitions with or without the rigid transformation flow. The first row shows renderings using our full model with both the RT and OT flows.
The second row shows renderings without the RT flow; only the OT flow is used. It can be seen that morphing without the RT flow directly moves each point toward the target and results in shattered rendering. Such an effect is unsatisfactory as the edge of the plate falls into pieces. In contrast, our method gradually transforms the plate so that the hotdog and the ficus have their poses aligned. During transformation, the OT flow simultaneously performs deformation in local regions so the texture can better resemble the target.
We also compare to the baseline, where no flows but simple blending is applied, as shown in the third row. Morphing with cross-dissolve leads to ghost effects.


\paragraph{Optimal transport flow:}
Our flow composition method generates smooth morphing by jointly performing rigid transformation and OT deformation. Here we examine the effect when only one of the two flows is used. This can be manipulated by the blending weight $t$ in each flow. As previously shown in \cref{fig:flow}, we use only the RT flow in the first half of morphing, and the OT flow in the second half. The source scene is on the left, and the subsequent three images are only affected by the RT flow. We can see that the flow aligns the poses of the source and target shapes. When the two scenes are fully aligned, the OT flow deforms the hotdog into the ficus, as shown in the fifth and sixth images.

\section{Conclusion}
This paper introduces Multiview Regenerative Morphing---a new method that integrates volume rendering and optimal transport to address a new task of multiview image morphing. Our method can produce interesting morphing effects that have not yet been demonstrated by previous image-based morphing methods. From the multiview images of two category-agnostic scenes without predefined correspondences, our method learns volumetric representations to render free-form morphs that can be visualized from arbitrary perspectives at any transition moment. We decouple the morphing process into two flows in Wasserstein metrics: one governs the rigid transformation and the other models the correspondences and deformations. The two flows estimated via optimization then jointly provide as-rigid-as-possible transformation under required topological and morphological changes between the two shapes. Our method is fast in training; it takes less than half an hour to learn the morphing renderer from both scenes, which otherwise might need 30x longer time if learned by typical neural rendering methods. The learned morphing renderer can readily generate on-the-fly multiview morphs showcasing new visual effects.

\clearpage

\begin{center}
    \Large\bfseries\boldmath
    Appendix
\end{center}

\appendix

\section{Sinkhorn Divergence and Morphing Flows}
\label{app_sec:intro}

Sinkhorn divergence \cite{genevay2018learning,feydy2019interpolating} is a popular metric for comparing distributions in the form of regularized optimal transport. Given two positive, unit-mass measures $\gamma,\beta \in \mathcal{M}^{+}(\mathcal{X})$ on some feature space $\mathcal{X}$, the Sinkhorn divergence with parameter $\varepsilon$ is expressed as
\begin{equation}
\label{eq: Sinkhorn}
    \Sinkhorn_{\varepsilon}(\gamma, \beta) = \mathrm{OT}_{\varepsilon}(\gamma, \beta)-\frac{1}{2} \mathrm{OT}_{\varepsilon}(\gamma, \gamma)-\frac{1}{2} \mathrm{OT}_{\varepsilon}(\beta, \beta) \, ,
\end{equation}
where $\mathrm{OT}_\varepsilon$ is an entropy-regularized optimal transport \cite{cuturi2013sinkhorn,peyre2019computational} that can be solved using the famous Sinkhorn algorithm. The optimal transport $\mathrm{OT}_\varepsilon$ is defined as
\begin{equation}
\label{eq: OT}
    \mathrm{OT}_{\varepsilon}(\gamma, \beta) = \min _{\pi_{1}=\gamma, \pi_{2}=\beta} \int_{\mathcal{X}^{2}} \mathrm{C} \,\mathrm{d} \boldsymbol{\pi}+\varepsilon \, \mathrm{KL}(\boldsymbol{\pi} \mid \gamma \otimes \beta) \, ,
\end{equation}
where $\mathrm{KL}$ is the Kullback-Leibler divergence. The coupled measure $\boldsymbol{\pi}\in \mathcal{M}^{+}(\mathcal{X}^2)$ has two marginals $(\pi_1, \pi_2)$, and the optimization is to find the the transport plan $\boldsymbol{\pi}$ that moves all the mass of $\gamma$ toward $\beta$, subject to the constraints on the two marginals that $\pi_1 = \gamma$ and $\pi_2 = \beta$. The distance function $\mathrm{C}$ is defined as $\mathrm{C}(x,y) = \|x-y\|^p$ and often referred to as the \emph{p--Wasserstein} distance.
\cref{eq: OT} can be optimized by taking the Sinkhorn iterations with its dual form. For discretely sampled measures, such as $\gamma=\sum_{i=1}^{\mathrm{N}} \bgamma_{i} \Delta_{\mathbf{x}_{i}}$ and $\beta=\sum_{j=1}^{\mathrm{M}} \bbeta_{j} \Delta_{\mathbf{y}_{j}}$, the optimized $\mathrm{OT}_{\varepsilon}$ can be expressed as
\begin{equation}
    \mathrm{OT}_{\varepsilon}\left(\{(\bgamma_{i}, \mathbf{x}_{i})\}, \{(\bbeta_{j}, \mathbf{y}_{j})\}\right)=\sum_{i=1}^{\mathrm{N}} \bgamma_{i} \mathbf{u}_{i}+\sum_{j=1}^{\mathrm{M}} \bbeta_{j} \mathbf{v}_{j} \, ,
\end{equation}
where $(\mathbf{u},\mathbf{v})$ are the optimal dual vectors with
\begin{equation}
\label{eq: dual vectors}
\begin{aligned}
    \mathbf{u}_{i} &= -\varepsilon \,\log \sum_{k=1}^M \exp \left(\log \left(\bbeta_{k}\right)+\frac{1}{\varepsilon} \mathbf{v}_{k}-\frac{1}{\varepsilon} \mathrm{C}(\mathbf{x}_{i}, \mathbf{y}_{k})\right) \, , \\
    \mathbf{v}_{j} &= -\varepsilon \,\log \sum_{k=1}^N \exp \left(\log \left(\bgamma_{k}\right)+\frac{1}{\varepsilon} \mathbf{u}_{k}-\frac{1}{\varepsilon} \mathrm{C}(\mathbf{x}_{k}, \mathbf{y}_{j})\right) \, .
\end{aligned}
\end{equation}
Therefore, the Sinkhorn divergence between $\gamma$ and $\beta$ in the discrete form can be computed as
\begin{equation}
\label{eq: Sinkhorn dual}
\begin{aligned}
    \Sinkhorn_\varepsilon  \left(\{(\bgamma_{i}, \mathbf{x}_{i})\}, \{(\bbeta_{j}, \mathbf{y}_{j})\}\right)
    = \sum_{i=1}^{\mathrm{N}} \bgamma_{i} \left(\mathbf{u}_{i}-\mathbf{a}_i\right) + \sum_{j=1}^{\mathrm{M}} \bbeta_{j} \left(\mathbf{v}_{j}-\mathbf{b}_j\right) \, ,
\end{aligned}
\end{equation}
where $\mathbf{a}_i$, $\mathbf{b}_j$ are the optimal dual vectors that yield optimal $\mathrm{OT}_{\varepsilon}(\gamma, \gamma)$ and $\mathrm{OT}_{\varepsilon}(\beta, \beta)$, the regularization terms in Sinkhorn divergence. 
    We can minimize Sinkhorn divergence by moving mass from $\bx_i$ to $\by_j$. This is achieved by taking the derivative $\partial_{\bx_i} \Sinkhorn_\varepsilon = \bgamma_i \nabla \left(\mathbf{u}_{i}-\mathbf{a}_i\right) = \bgamma_i \nabla \Phi(\mathbf{x}_i)$, where $\Phi$ is defined as
\begin{equation}
\label{eq: Phi_x}
\begin{aligned}
\Phi(\mathbf{x})=&-\varepsilon \log \sum_{j=1}^{M} \exp \left[\log \left(\bbeta_j \right)+\frac{1}{\varepsilon} \mathbf{v}_{j}-\frac{1}{\varepsilon} \mathrm{C}(\mathbf{x}, \by_{j})\right] \\
&+\varepsilon \log \sum_{i=1}^{N} \exp \left[\log \left(\bgamma_i \right)+\frac{1}{\varepsilon} \mathbf{a}_{i}-\frac{1}{\varepsilon} \mathrm{C}(\mathbf{x}, \mathbf{x}_{i})\right] \, .
\end{aligned}
\end{equation}
Based on \cref{eq: dual vectors,eq: Sinkhorn dual}, if we plug $\bx_i$ into \cref{eq: Phi_x} we can get $\Phi(\bx_i) = \bu_i-\ba_i$ as expected.


In our experiment, the source and target shape representations $\cS$ and $\cT$ are the collections of their opacity information and the coordinates of the respective voxels. The representations can be expressed as sums of weighted Dirac masses
\begin{equation}
    \cS = \sum_{i=1}^{N^\cS} \omega_i^\cS  \, \Delta_{\bx_i^\cS}, \quad  \cT = \sum_{j=1}^{N^\cT} \omega_j^\cT  \, \Delta_{\bx_j^\cT} \, ,
\end{equation}
where $\omega_i^\cS$ and $\omega_j^\cT$ represent the normalized opacity collections $\alpha$ derived from $\cV_\alpha^\cS$ and $\cV_\alpha^\cT$. We aim to minimize Sinkhorn divergence $\Sinkhorn_\varepsilon \left(\{(\omega_{i}^\cS, \mathbf{x}_{i}^\cS)\}, \{(\omega_{j}^\cT, \mathbf{x}_{j}^\cT)\}\right)$ between $\cS$ and $\cT$ using the rigid transformation flow and the optimal transport flow.

\subsection{Rigid transformation flow}
The rigid transformation flow produces an as-rigid-as-possible visual effect for morphing.
Without a definite pose between two unrelated objects, the flow estimates an optimal 6D transformation $\RT \in \mathrm{SE(3)}$, with rotation $\mathbf{R}$ and translation $\bz$, that best aligns the shapes of the two objects. The optimization can be formulated as
\begin{equation}
    \hat{\RT} = \argmin_\RT \; \Sinkhorn_\varepsilon \left(\left\{\left(\omega_{i}^\cS, \RT(\mathbf{x}_{i}^S)\right)\right\}, \{(\omega_{j}^\cT, \mathbf{x}_{j}^\cT)\}\right) \,,
\end{equation}
which is optimized through gradients descent. Considering \cref{eq: Sinkhorn dual}, the differentiation of $\Sinkhorn_\varepsilon$ with respect to $\RT$ is
\begin{equation}
\label{eq: Sinkhorn diff}
\begin{aligned}
    \frac{1}{\omega_{i}^\cS} \partial_{\RT} \Sinkhorn_{\varepsilon} \left(\left\{\left(\omega_{i}^\cS, \RT(\bx_{i}^\cS)\right)\right\}, \left\{\left(\omega_{j}^\cT, \bx_{j}^\cT \right)\right\} \right)
    =\nabla \Phi\left(\RT(\bx_{i}^\cS)\right) \cdot \frac{\partial \RT(\bx_{i}^\cS)}{\partial \RT} \, ,
\end{aligned}
\end{equation}
where $\Phi(\cdot)$ is defined in \cref{eq: Phi_x} with specialization on $\left\{\left(\omega_{i}^\cS, \RT(\bx_{i}^\cS)\right)\right\}$ and $\left\{\left(\omega_{j}^\cT, \bx_{j}^\cT \right)\right\}$.
Using the above equations, we can update every component of $\RT$. In our experiment, we use \emph{GeomLoss} \cite{feydy2019interpolating} to calculate the Sinkhorn divergence, and \emph{Pytorch} to compute the gradient and the partial differentiation in \cref{eq: Sinkhorn diff}. Since the update on the rotation matrix $\mathbf{R}$ does not guarantee to be a rotation matrix, we use singular value decomposition to normalize the updated $\mathbf{R}$ in every iteration. After $K$ iterations, we obtain an estimate $\hat{\RT}$ and the rigid transformation flow $\mathbf{f}$:
\begin{equation}
\label{eq: transformation flow}
    \mathbf{f}(t) = t \cdot (\hat{\RT}(\cS) -  \cS) \,.
\end{equation}
$\hat{\RT}(\cS)$ is also used for computing the optimal transport flow.

\subsection{Optimal transport flow}
The optimal transport flow deforms the source shape to the target shape using the gradient of Sinkhorn divergence. Consider a Dirac mass located at $\bx_i$, with weight $\omega_i^\cS$ and belonging to the transformed source shape $\hat{\RT}(\cS)$, the optimal transport flow related to $\bx_i$ can be written as
\begin{equation}
\label{eq: barycenter flow}
    \mathbf{g}_i(t) = -t \cdot  \partial_{\mathbf{x}_i} \Sinkhorn(\hat{\RT}(\cS), \cT)  \,,
\end{equation}
for some blending weight $t \in [0,1]$. In our experiment, the flow deforms from rigidly-transformed source shape to target shape. The setting constrains the deformation to happen between aligned objects, instead of deforming directly from source to target that leads to a shattered shape.
Therefore, we can formulate \cref{eq: barycenter flow} as
\begin{equation}
\begin{aligned}
    \mathbf{g}_i(t) &= -t \cdot \partial_{\bx_i} \Sinkhorn_{\varepsilon} \left( \left\{\left(\omega_{i}^\cS, \hat{\RT} (\bx_{i}^\cS)\right)\right\}, \left\{\left(\omega_{j}^\cT, \bx_{j}^\cT \right)\right\} \right) \\
    &= -t \cdot \nabla \Phi\left(\hat{\RT}(\bx_{i}^\cS)\right) \, ,
\end{aligned}
\end{equation}
which can be computed by \cref{eq: Phi_x} with specialization on $\left\{\left(\omega_{i}^\cS, \hat{\RT}(\bx_{i}^\cS)\right)\right\}$ and $\left\{\left(\omega_{j}^\cT, \bx_{j}^\cT \right)\right\}$. The flow component $\mathbf{g}_i$ moves $\bx_i^\cS$ from $\hat{\RT}(\cS)$ to close to $\cT$.

\section{Morphing Results}
\begin{figure*}
    \centering
    \includegraphics[width=0.99\textwidth]{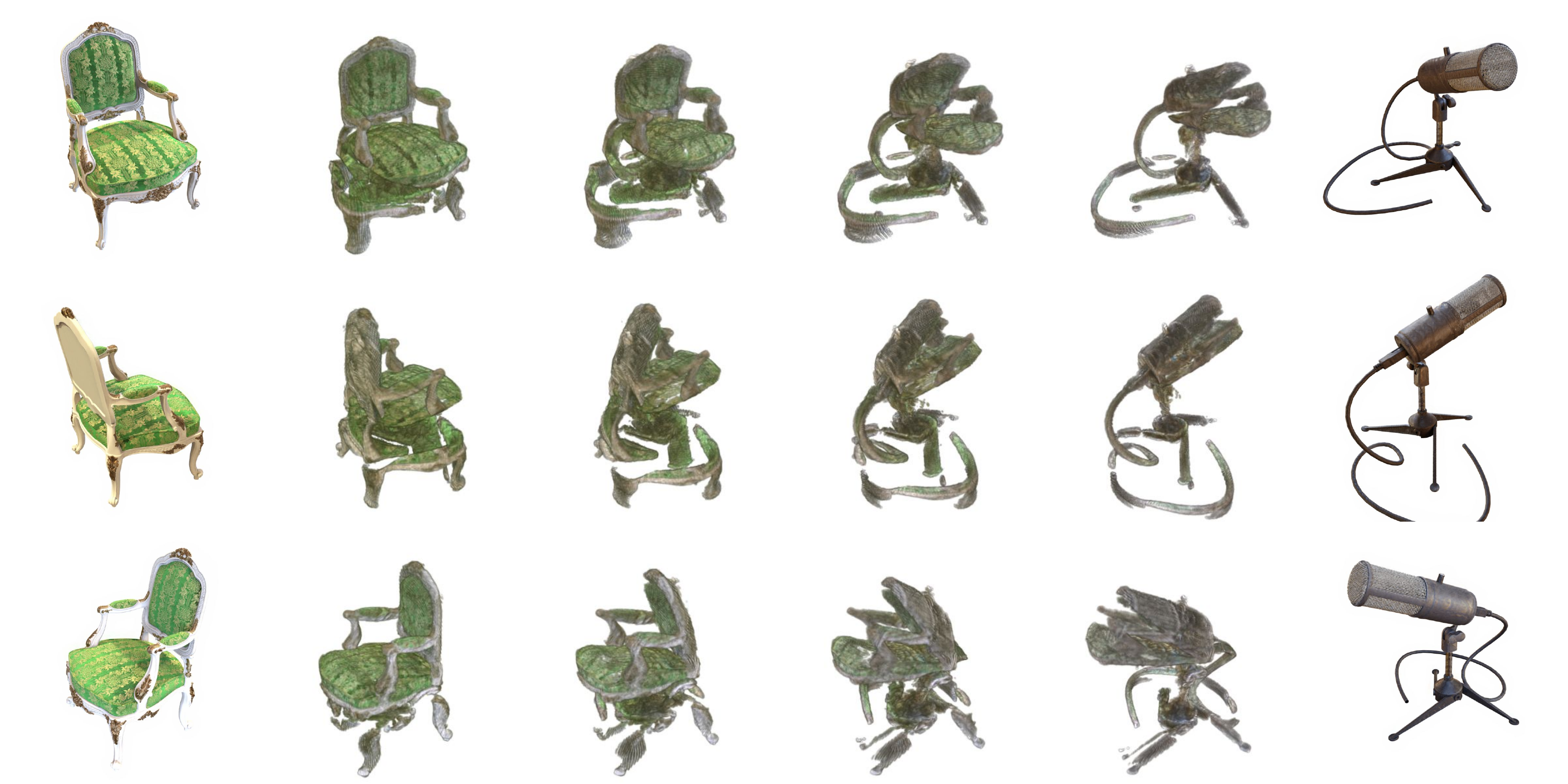}
    \caption{The morphing shows the transition from `Chair' to `Mic'. From left to right, the morphs have blending weights $t = \{0, .2, .4, .6, .8, 1\}$. From top to down, the images are rendered with azimuth $\theta = \{30\degree, 150\degree, 270\degree\}$ and elevation $\phi = 30\degree$. Note that the weight $t$, the azimuth $\theta$, and the elevation $\phi$ are chosen randomly and can be replaced with different values.}
    \label{fig:chair2mic}
\end{figure*}

\begin{figure*}
    \centering
    \includegraphics[width=0.99\textwidth]{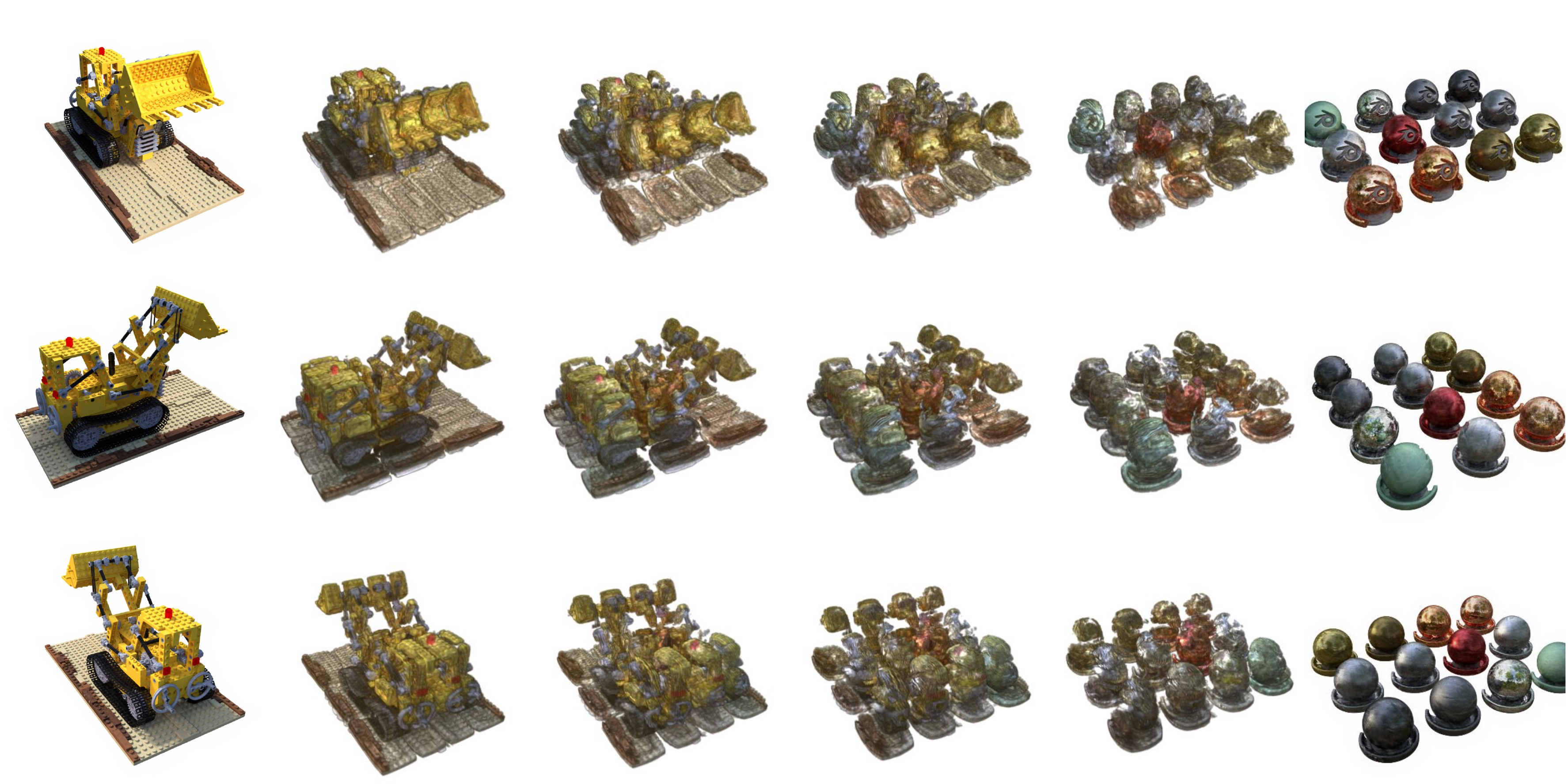}
    \caption{The morphing shows the transition from `Lego' to `Materials'. From left to right, the morphs have blending weights $t = \{0, .2, .4, .6, .8, 1\}$. From top to down, the images are rendered with azimuth $\theta = \{30\degree, 150\degree, 270\degree\}$ and elevation $\phi = 30\degree$. Note that the weight $t$, the azimuth $\theta$, and the elevation $\phi$ are chosen randomly and can be replaced with different values.}
    \label{fig:lego2mat}
\end{figure*}

\begin{figure*}
    \centering
    \includegraphics[width=0.97\textwidth]{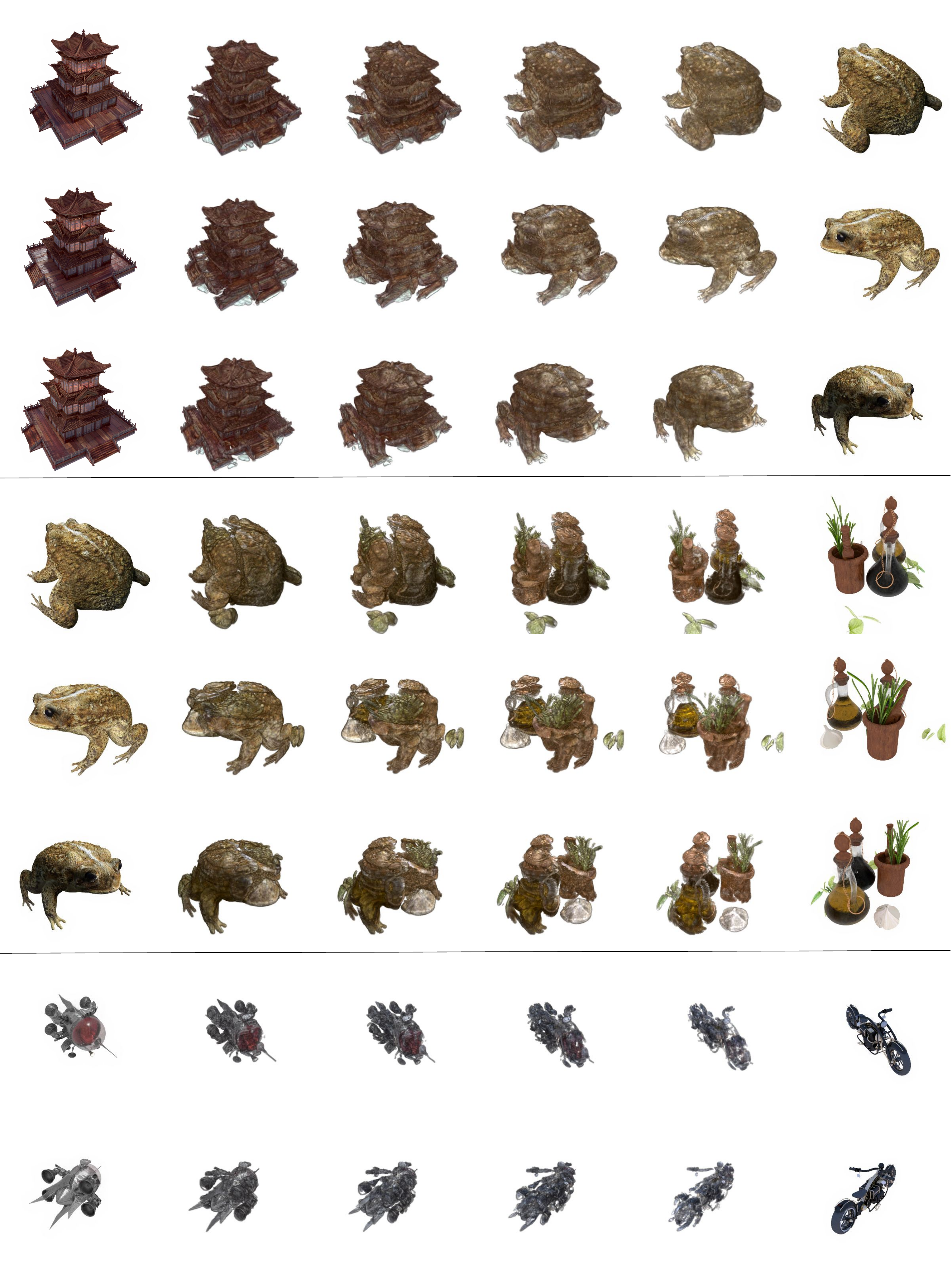}
    \caption{Rendering in more views of morphing results of Synthetic-NSVF shown in the main paper. Each row presents the morphing process as in \cref{fig:chair2mic}, but with different azimuths. In the first and the second canvases, the rows are rendered with azimuth $\theta = \{30\degree, 150\degree, 270\degree\}$, from top to bottom. While in the third canvas, the rows are rendered with $\theta = 60\degree$ and $240\degree$.
    }
    \label{fig:nsvf_view}
\end{figure*}

\begin{figure*}
    \centering
    \includegraphics[width=0.97\textwidth]{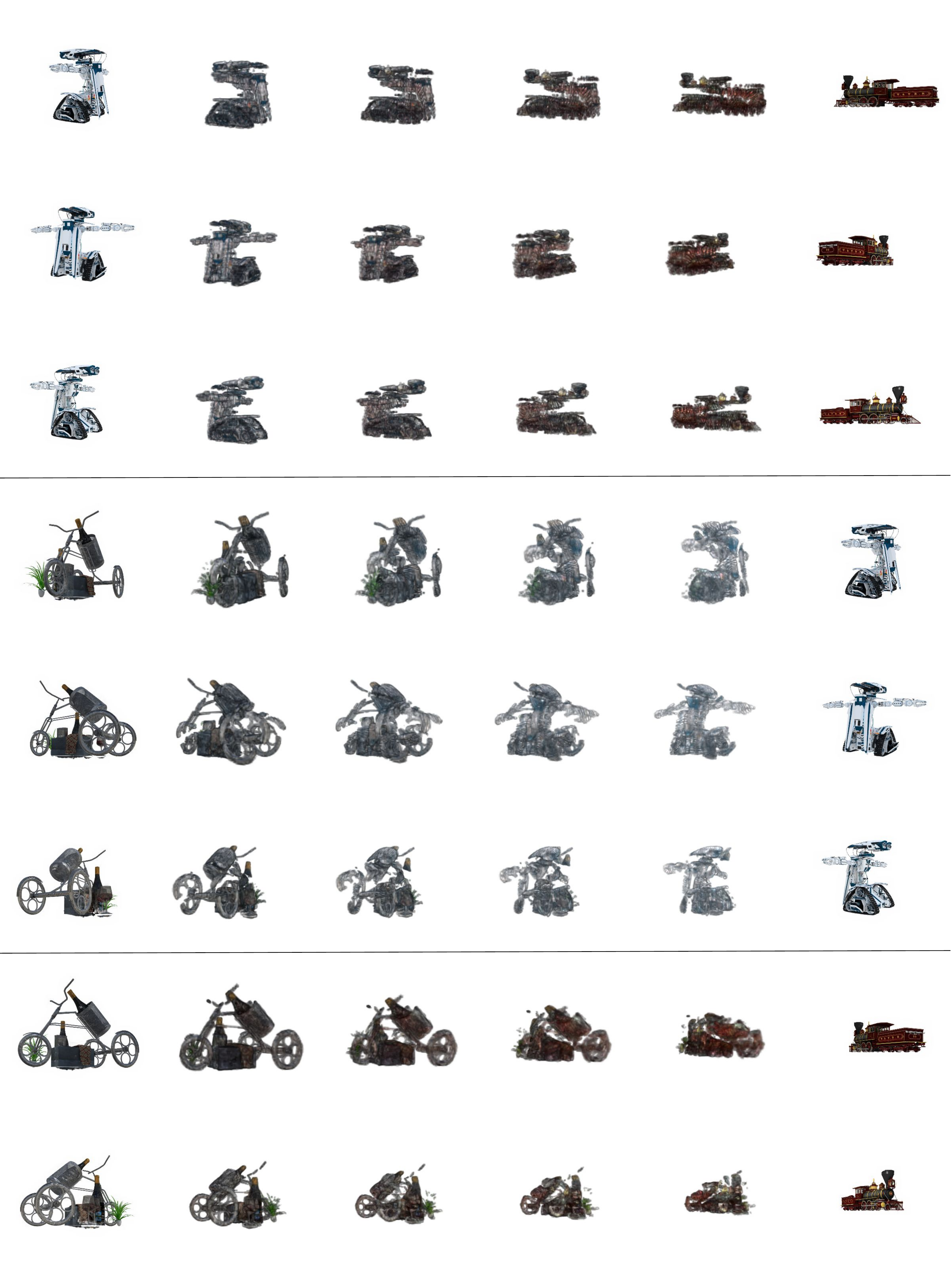}
    \caption{More morphing results between scenes from Synthetic-NSVF. Each row presents the morphing process as in \cref{fig:chair2mic}, but with different azimuths. In the first and the second canvases, the rows are rendered with azimuth $\theta = \{30\degree, 150\degree, 270\degree\}$, from top to bottom. While in the third canvas, the rows are rendered with $\theta = 60\degree$ and $240\degree$.
    }
    \label{fig:nsvf_other}
\end{figure*}

We show in \cref{fig:chair2mic,fig:lego2mat,fig:nsvf_view,fig:nsvf_other} more morphing results on different combinations of sources and targets rendered in different viewing angles. For more results and multiview-consistent visualizations please see the video. \cref{fig:chair2mic} shows the transition from `Chair' to `Mic', rendered in three different views. \cref{fig:lego2mat} shows morphs between `Lego' and `Materials' in three different views. \cref{fig:nsvf_view} renders more views of morphing with Synthetic--NSVF shown in the main paper. \cref{fig:nsvf_other} demonstrates more morphing results of scenes from Synthetic--NSVF.

\ignore{
\begin{equation}
\label{eq: Sinkhorn dual}
\begin{aligned}
    \Sinkhorn_\varepsilon & \left(\{(\bgamma_{i}, \mathbf{x}_{i})\}, \{(\bbeta_{j}, \mathbf{y}_{j})\}\right) \\
    &= \sum_{i=1}^{\mathrm{N}} \bgamma_{i} \left(\mathbf{u}_{i}-\mathbf{a}_i\right) + \sum_{j=1}^{\mathrm{M}} \bbeta_{j} \left(\mathbf{v}_{j}-\mathbf{b}_j\right) \\
    \partial_{\bx_i} \Sinkhorn_\varepsilon &= \bgamma_i \nabla \left(\mathbf{u}_{i}-\mathbf{a}_i\right) = \nabla \Phi(\mathbf{x}_i) \\
\end{aligned}
\end{equation}
$\Phi(\bx)$ is defined in Eq. (9). In Eq. (8), instead of $\partial_{\bx_i} \Sinkhorn_\varepsilon$, we want to find $\partial_{\bW \bx_i} \Sinkhorn_\varepsilon$ and $\partial_{\bW} \bW \bx_i$, so we can update parameter $\bW$, phrased here as $\RT$
}


%
%

\bibliographystyle{splncs04}
\bibliography{nerf,morph,ot}
\end{document}